\documentclass{IOS-Book-Article}

\usepackage{mathptmx}

\usepackage{tabularx}
\usepackage{rotating}
\usepackage{booktabs}
\usepackage{multirow}
\newcommand{\PreserveBackslash}[1]{\let\temp=\\#1\let\\=\temp}

\def\hb{\hbox to 10.7 cm{}}

\begin{document}

\pagestyle{headings}
\def\thepage{}

\begin{frontmatter}              

\title{Better Multi-class Probability Estimates for Small Data Sets}

\markboth{}{January 2020\hb}

\author[A]{\fnms{Tuomo} \snm{Alasalmi}%
\thanks{Corresponding Author: Tuomo Alasalmi, Biomimetics and Intelligent Systems Group, University of Oulu, P.O. Box 4500, FI-90014 University of Oulu, Finland. E-mail: tuomo.alasalmi@oulu.fi}},
\author[A]{\fnms{Jaakko} \snm{Suutala}},
\author[B]{\fnms{Heli} \snm{Koskim{\"a}ki}},
and
\author[A]{\fnms{Juha} \snm{R{\"o}ning}}

\runningauthor{T. Alasalmi et al.}
\address[A]{Biomimetics and Intelligent Systems Group, University of Oulu, Finland.}
\address[B]{Oura Health Ltd., Oulu, Finland.}

\begin{abstract}
Many classification applications require accurate probability estimates in addition to good class separation but often classifiers are designed focusing only on the latter. Calibration is the process of improving probability estimates by post-processing but commonly used calibration algorithms work poorly on small data sets and assume the classification task to be binary. Both of these restrictions limit their real-world applicability. Previously introduced Data Generation and Grouping algorithm alleviates the problem posed by small data sets and in this article, we will demonstrate that its application to multi-class problems is also possible which solves the other limitation. Our experiments show that calibration error can be decreased using the proposed approach and the additional computational cost is acceptable.
\end{abstract}

\end{frontmatter}
\markboth{January 2020\hb}{January 2020\hb}

\section{Introduction}
\label{sec:introduction}

In classification, an object of interest is predicted to belong to one of discrete and predefined categories called classes. An example of a classification problem would be recognizing handwritten digits. In many applications it is also important to quantify the uncertainty of these predictions. In the handwritten digits example, how certain can we be that the digit is one and not seven or any other digit? If the results of a classifier are used as input for making decisions or if there are costs involved in the classification decision, then it is important, in addition to good classification accuracy, that the probabilities predicted by a classifier are accurate. A classifier is said to be well calibrated if the predicted probability of an event is close to the proportion of these events among a group of similar predictions \cite{Dawid1982}. However, the main objective for classifier design is often good class separation and not accurate probability estimation. Therefore, many commonly used classifiers are not well calibrated. The process for improving a classifiers probability estimates by post-processing the probability estimates is called calibration. Most commonly used calibration algorithms only work on binary problems and need a fair amount of data, separate from training and testing data to avoid bias, which severely restricts their application in real-world problems.

To tackle these two limitations, we will demonstrate two ways to generalize a binary calibration method that has been previously shown to work on small data sets to work on multi-class problems. Using the proposed calibration approach lead to statistically significant improvement in calibration error metrics. The rest of this article is structured as follows. Section \ref{sec:background} will shortly review relevant literature on the topic, Section \ref{sec:experiments} explains the experiments that were used for testing the proposed approaches and results from those experiments are presented in Section \ref{sec:results}. The results are discussed in Section \ref{sec:discussion} and Section \ref{sec:conclusions} concludes the article.

\section{Background}
\label{sec:background}

Calibration algorithms need training data and to avoid biasing this data needs to be separate from the data that is used for training the classifier. A minimum of about 1000 to 2000 training samples are needed for the calibration data set depending on the learning algorithm to avoid overfitting. Non-parametric calibration algorithms are particularly prone to overfitting on small data sets and their performance seems to improve with increasing calibration data set sizes even further \cite{NiculescuMizil2005ICML}. This means that the amount of training data in total needs to be large so that enough data can be set aside for calibration. In addition, a separate data set needs to be held out for testing. However, relatively small data sets are quite common in many real-world modelling tasks.

It has been previously shown that calibrating binary classifiers with traditional calibration approach does not work very well when available data is limited. However, it is possible to solve the problem, at least partially, by generating more calibration data with a Monte Carlo cross validation approach \cite{Alasalmi2018,Alasalmi2020} using isotonic regression (IR) \cite{NiculescuMizil2005ICML} or ensemble of near isotonic regression models (ENIR) \cite{Naeini2018} calibration algorithms. Many classification problems are not binary but instead the problem often is to classify the data into multiple classes ($K > 2$) but most calibration algorithms work on binary ($K = 2$) classification problems only. This is also true for the above mentioned solution that uses Data Generation and Grouping (DGG) algorithm \cite{Alasalmi2020} which works with binary calibration algorithms only.

A solution to this problem is to break the multi-class problem into several binary problems, solve each binary classification and calibration problem independently, and combine the results to multi-class probability estimates \cite{Zadrozny2002}. The premise is obviously that better calibrated binary probabilities result in better calibrated multi-class probabilities. The question then becomes how to divide the problem into binary problems and how to combine the results. Two intuitive ways to break the multi-class problem into binary problems are one-vs-rest and all pairs.

In the one-vs-rest approach the binary problems are such that one of the classes is treated as the positive class while the rest are treated as the negative class collectively and this is repeated for each class. The number of binary problems then becomes the same as the number of classes $K$. Probability estimates from using the one-vs-rest approach can be combined by simply linearly normalizing the binary probabilities for each class so that they sum up to one. This results in comparable error rates with combining the probabilities using least squares or coupling algorithms \cite{Zadrozny2002}. By using one class as the positive class and the rest of the data as the negative class leads to class imbalance which becomes more pronounced as the number classes grows. However, the number of binary problems in this approach remains reasonable.

In the all pairs approach all possible pairs of classes are enumerated and one class in each pair is selected as the positive class while the other class serves as the negative class. There are $K(K-1)/2$ possible pairs of classes in this approach meaning that the number of binary problems is larger than with the one-vs-rest approach when $K > 3$ as can be seen from Table \ref{tab:bin_problems}. However, the binary problems are faster to learn in all pairs approach as only instances from the two classes are included in each. The binary problems are also more balanced in the all pairs approach. After learning and calibrating the binary classifiers, the probabilities for the multi-class problem can be combined with pairwise coupling which was originally developed by Hastie and Tibshirani \cite{Hastie1998} and later improved by Wu et al. \cite{Wu2004}.

\begin{table}[t]
\centering
\caption{The number of binary problems for One-vs-rest and All Pairs approaches.}
\label{tab:bin_problems}       
\begin{tabular}{ccc}
\toprule
K & One-vs-rest & All Pairs  \\
\midrule
3 & 3 & 3 \\
4 & 4 & 6 \\
5 & 5 & 10 \\
6 & 6 & 15 \\
... & ... & ... \\
10 & 10 & 45 \\
\bottomrule
\end{tabular}
\end{table}

The two above mentioned intuitive ways for breaking up the multi-class problem are two special cases of a more general idea that uses so called error correcting output coding (ECOC) matrices \cite{Allwein2000}. ECOC matrices can be either complete or sparse. However, the number of binary problems grows exponentially as the number of classes grows when using complete ECOC matrices and there are computational problems with sparse ECOC matrices making both infeasible in practice \cite{Gebel2009}.

\section{Experiments}
\label{sec:experiments}

In this study, the feasibility of the DGG data generation algorithm for multi-class classification problem calibration was tested. One-vs-rest approach with normalization and all pairs approach with pairwise coupling were compared here when using the DGG algorithm along with ENIR calibration. The procedure in the context of binary calibration is described more thoroughly in \cite{Alasalmi2020}. Calibration error was quantified with logarithmic loss (LL) and mean squared error (MSE). LL is defined in Equation \ref{eq:logloss} and MSE in Equation \ref{eq:mse}. In the equations $N$ stands for the number of observations, $K$ stands for the number of class labels, $log$ is the natural logarithm, $y_{i,j}$ equals $1$ if observation $i$ belongs to class $j$, otherwise it is $0$, and $p_{i,j}$ stands for the predicted probability that observation $i$ belongs to class $j$. A smaller value of each metric indicates better calibration.

\begin{equation}
LL = -\frac{1}{N} \sum_{i=1}^{N} \sum_{j=1}^{K} y_{i,j} log(p_{i,j})
\label{eq:logloss}
\end{equation}

\begin{equation}
MSE = \frac{1}{N} \sum_{i=1}^{N} \sum_{j=1}^{K} (y_{i,j} - p_{i,j})^2
\label{eq:mse}
\end{equation}

A stratified 10-fold cross validation was used to create data samples and Student's paired t-test with unequal variance assumption and the Welch modification to the degrees of freedom \cite{welch1947} was used to determine if there was a statistically significant difference between calibration scenarios.

Properties of the data sets that were used in the experiments are presented in Table \ref{tab:data_sets}. With the Abalone data set the task is to predict the age groups of abalones based on some physical measurements \cite{nash1994population}. Many of the classes had only a handful, some just one sample so classes 1 to 5 were grouped together as were groups 14 and 15, and all classes over 16. Contraceptive Method Choice data set (Contraceptive) is a subset of the 1987 National Indonesia Contraceptive Prevalence Survey. The task with the Contraceptive data is to predict the choice of current contraceptive method: long term, short term, or no use. The development index (development) data set describes the development status of countries based on demographic data and the task is to predict the development index of each country. The ecoli data set describes protein localization sites in Escherichia coli bacteria \cite{nakai1992knowledge}. Due to small number of samples classes were grouped so that sub classes of inner membrane were grouped together as were sub classes of outer membrane. The forest type mapping (forest) data set describes forested areas in Japan based on multi-temporal remote sensing data \cite{johnson2012using} and the task is to discriminate different forest types. Heart disease Cleveland data set (Heart) contains clinical and noninvasive test results of patients undergoing angiography at the Cleveland Clink \cite{Detrano1989}. Six patients with missing values were discarded from the analysis. The goal with the Heart data is to predict the severity of heart disease based on the patient data. The optical recognition of handwritten digits (pendigits) data set \cite{kaynak1995methods} contains preprocessed features that describe handwritten digits and the classification task is to recognize the digits. To facilitate the comparison of algorithm performance, the original division of training and test data sets was not used and instead the data sets were combined and cross validation was used like with the rest of the data sets. The seeds data set describes different varieties of wheat seeds based on a soft X-ray technique \cite{charytanowicz2010complete} and the task is to classify the seeds into correct class. The steel plates faults (steel) data set\footnote{Dataset provided by Semeion, Research Center of Sciences of Communication, Via Sersale 117, 00128, Rome, Italy. www.semeion.it} describes faults in steel plates and the task is to classify the plate faults into correct categories based on measurement data. Waveform data set consists of artificially generated data on three classes of waves described by noisy attributes and the classification task is to separate the wave classes \cite{Breiman1984}. The wholesale customers (wholesale) data set contains data of customers of a wholesale distributor in Portugal \cite{abreu2011analise}. The task is to classify the customer belonging into a certain region. The yeast data set describes protein localization sites with results from analysis techniques \cite{horton1996probabilistic} and the task is to classify each protein to the correct location based on the analysis results. Development index data set is available from kaggle data sets. Rest of the data sets used in the experiments are freely available from the UCI machine learning repository \cite{Lichman:2013}.

\begin{table}
\centering
\caption{Data set properties.}
\label{tab:data_sets}
\begin{tabular}{lccccc}
\toprule
Data set & Samples & Classes & Smallest & Largest \\
\midrule
Abalone & 4177 & 11 & 189 & 689 \\
Contraceptive & 1473 & 3 & 333 & 629 \\
Development & 212 & 4 & 13 & 89 \\
Ecoli & 336 & 4 & 25 & 143 \\
Forest & 523 & 4 & 83 & 195 \\
Heart & 297 & 5 & 13 & 160 \\
Pendigits & 10992 & 10 & 1055 & 1144 \\
Seeds & 210 & 3 & 70 & 70 \\
Steel & 1941 & 7 & 55 & 673 \\
Waveform & 5000 & 3 & 1647 & 1696 \\
Wholesale & 440 & 3 & 47 & 316 \\
Yeast & 1479 & 9 & 20 & 463 \\
\bottomrule
\end{tabular}
\end{table}

DGG data generation with ENIR calibration has been shown to work well especially with naive Bayes (NB) and random forest (RF) classifiers on binary problems and they are both capable of producing multi-class probability estimates without modification so they were selected as the base classifiers for our experiments. A total of five different calibration scenarios were compared in this study: multi-class uncalibrated probabilities (Multi-class Raw), one-vs-rest with either uncalibrated (One-vs-rest Raw) or calibrated (One-vs-rest DGG + ENIR) probabilities, and all pairs with either uncalibrated (All pairs Raw) or calibrated (All pairs DGG + ENIR) probabilities. In addition to calibration error metrics, computation times were recorded on a computational server (Intel Xeon E5-2650 v2 @ 2.60GHz, 196GB RAM) for each calibration scenario.

\section{Results}
\label{sec:results}

Results of the experiments are summarized in Table \ref{tab:results_summary} which shows how many of the data sets had statistically significant changes in calibration performance after our calibration treatment on the data sets grouped by the classifier used, the approach to form the binary problems, and by the number of classes. Full results, MSEs and LLs, for each of the tested data sets in each calibration scenario are reported in Tables \ref{tab:results_summary_nb} and \ref{tab:results_summary_rf} for naive Bayes and random forest, respectively.

\begin{table}
\centering
\caption{Effect of the binary problem division method to calibration performance on the 12 data sets. Problems with three to five classes are considered to have low number of classes (8 pcs) and problems with seven or more classes as having high number of classes (4 pcs).}
\label{tab:results_summary}
\begin{tabular}{lccl}
\toprule
\multirow{2}{*}{Classifier}
    & \multicolumn{2}{c}{Number of classes} & \\
    & Low & High & \\
\midrule
\multirow{6}{*}{NB}
    & $6 \times \uparrow$ & $4 \times \uparrow$ & \\
    & $2 \times \leftrightarrow$ & & One-vs-rest \\
    & & & \\
    \cmidrule{2-4}
    & $5 \times \uparrow$ & $2 \times \uparrow$ & \\
    & $3 \times \leftrightarrow$ & $1 \times \leftrightarrow$ & All Pairs \\
    & & $1 \times \downarrow$ & \\
\midrule
\multirow{6}{*}{RF}
    & $3 \times \uparrow$ & $1 \times \uparrow$ & \\
    & $5 \times \leftrightarrow$ & $3 \times \leftrightarrow$ & One-vs-rest \\
    & & & \\
    \cmidrule{2-4}
    & $3 \times \uparrow$ & & \\
    & $2 \times \leftrightarrow$ & & All Pairs \\
    & $3 \times \downarrow$ & $4 \times \downarrow$ & \\
\bottomrule
\addlinespace[\belowrulesep]
\multicolumn{4}{p{0.46\linewidth}}{\footnotesize{$\uparrow$ indicates improved calibration, $\leftrightarrow$ indicates neutral effect, and $\downarrow$ indicates impaired calibration.}}
\end{tabular}
\end{table}

\begin{table}
\centering
\caption{Mean squared error and logarithmic loss of naive Bayes classifier on different calibration scenarios.}
\label{tab:results_summary_nb}       
\begin{tabularx}{\linewidth}{p{1.55cm}XXXXXXXXXX}
\toprule
Data set & \multicolumn{2}{p{0.13\textwidth}}{Multi-class Raw} & \multicolumn{2}{p{0.13\textwidth}}{One-vs-rest Raw} & \multicolumn{2}{p{0.13\textwidth}}{One-vs-rest DGG+ENIR} & \multicolumn{2}{p{0.13\textwidth}}{All pairs Raw} & \multicolumn{2}{p{0.13\textwidth}}{\PreserveBackslash\raggedright All pairs DGG+ENIR} \\
\midrule
& MSE & LL & MSE & LL & MSE & LL & MSE & LL & MSE & LL \\ 
\midrule
Abalone & 0.089 & 4.676 & \underline{0.079} & \underline{3.634} & \textbf{\underline{0.074}} & \textbf{\underline{2.833}} & 0.169 & \underline{7.791} & \underline{0.079} & \underline{3.087} \\
Contraceptive & 0.233 & 2.354 & \underline{0.221} & \underline{2.160} & \underline{0.200} & \underline{1.752} & 0.233 & 2.354 & \textbf{\underline{0.199}} & \textbf{\underline{1.750}} \\
Development & 0.081 & 3.039 & 0.082 & 2.773 & 0.065 & \underline{1.038} & 0.090 & 3.104 & \textbf{\underline{0.060}} & \textbf{\underline{0.876}} \\
Ecoli & \textbf{0.029} & 0.692 & 0.032 & 0.692 & 0.033 & 0.688 & \underline{0.036} & 0.858 & \underline{0.032} & \textbf{0.594} \\
Forest & 0.065 & 3.384 & 0.076 & 2.066 & \textbf{0.059} & \textbf{\underline{0.996}} & \underline{0.128} & 3.572 & \underline{0.117} & \underline{1.529} \\
Heart & 0.130 & 2.549 & \underline{0.108} & \underline{2.096} & \underline{0.099} & \underline{1.635} & 0.123 & 2.487 & \textbf{\underline{0.098}} & \textbf{\underline{1.598}} \\
Pendigits & 0.027 & 2.077 & \underline{0.035} & \underline{1.597} & \textbf{\underline{0.024}} & \textbf{\underline{0.918}} & \underline{0.059} & \underline{2.303} & \underline{0.056} & \underline{1.909} \\
Seeds & 0.054 & 0.848 & 0.047 & 0.610 & \textbf{0.046} & \textbf{0.450} & 0.054 & 0.848 & 0.050 & 0.456 \\
Steel & 0.102 & 6.101 & \underline{0.084} & \underline{3.085} & \textbf{\underline{0.066}} & \textbf{\underline{1.515}} & - & - & \underline{0.088} & \underline{2.224} \\
Waveform & 0.109 & 1.545 & \underline{0.079} & \underline{0.717} & \underline{0.075} & \underline{0.725} & 0.109 & 1.545 & \textbf{\underline{0.070}} & \textbf{\underline{0.710}} \\
Wholesale & 0.207 & 2.612 & \underline{0.199} & \underline{2.294} & \textbf{\underline{0.148}} & \textbf{\underline{1.405}} & 0.207 & 2.612 & \textbf{\underline{0.148}} & \textbf{\underline{1.405}} \\
Yeast & 0.064 & 2.118 & 0.064 & \underline{2.039} & \textbf{\underline{0.063}} & \textbf{\underline{1.935}} & \underline{0.105} & \underline{4.510} & \underline{0.116} & \underline{4.248} \\
\bottomrule
\addlinespace[\belowrulesep]
\multicolumn{11}{p{0.966\textwidth}}{\footnotesize{Average results of 10-fold cross validation. Significantly different from Multi-class Raw is indicated with underlining. Best performing scenario with each classifier is indicated with boldface font.}}
\end{tabularx}
\end{table}

\begin{table}
\centering
\caption{Mean squared error and logarithmic loss of random forest classifier on different calibration scenarios.}
\label{tab:results_summary_rf}
\begin{tabularx}{\linewidth}{p{1.55cm}XXXXXXXXXX}
\toprule
Data set & \multicolumn{2}{  p{0.13\textwidth}}{Multi-class Raw} & \multicolumn{2}{  p{0.13\textwidth}}{One-vs-rest Raw} & \multicolumn{2}{  p{0.13\textwidth}}{One-vs-rest DGG+ENIR} & \multicolumn{2}{  p{0.13\textwidth}}{All pairs Raw} & \multicolumn{2}{  p{0.13\textwidth}}{\PreserveBackslash\raggedright All pairs DGG+ENIR} \\
\midrule
& MSE & LL & MSE & LL & MSE & LL & MSE & LL & MSE & LL \\ 
\midrule
Abalone & 0.073 & \textbf{2.859} & 0.073 & \underline{2.98}1 & \textbf{\underline{0.072}} & \underline{2.919} & \underline{0.080} & \underline{3.534} & \underline{0.078} & \underline{3.158} \\
Contraceptive & 0.186 & 1.682 & \underline{0.190} & \underline{1.752} & 0.184 & 1.661 & 0.185 & 1.652 & \textbf{\underline{0.181}} & \textbf{\underline{1.613}} \\
Development & 0.004 & 0.087 & \underline{0.009} & \underline{0.203} & \textbf{\underline{0.003}} & \textbf{\underline{0.057}} & \underline{0.022} & \underline{0.386} & \underline{0.011} & 0.730 \\
Ecoli & 0.031 & 0.685 & 0.029 & 0.559 & \textbf{0.028} & \textbf{0.442} & \underline{0.038} & 0.584 & 0.029 & 0.743 \\
Forest & 0.044 & \textbf{0.640} & 0.044 & 0.840 & \textbf{0.043} & 0.808 & \underline{0.106} & \underline{1.356} & \underline{0.108} & \underline{1.371} \\
Heart & 0.101 & 1.607 & 0.102 & 1.640 & \textbf{0.097} & \textbf{1.573} & 0.100 & 1.602 & 0.099 & 1.776 \\
Pendigits & 0.003 & 0.157 & \underline{0.003} & \underline{0.174} & \textbf{\underline{0.002}} & \textbf{\underline{0.105}} & \underline{0.055} & \underline{1.957} & \underline{0.053} & \underline{1.899} \\
Seeds & \textbf{0.033} & 0.359 & 0.035 & \textbf{0.358} & 0.037 & 0.371 & \underline{0.035} & 0.360 & \underline{0.037} & 0.678 \\
Steel & 0.041 & 1.005 & 0.041 & \textbf{1.002} & \textbf{\underline{0.039}} & 1.034 & - & - & \underline{0.094} & \underline{2.065} \\
Waveform & 0.076 & 0.759 & 0.075 & 0.749 & \textbf{\underline{0.066}} & \textbf{\underline{0.640}} & 0.076 & 0.755 & \underline{0.068} & \underline{0.680} \\
Wholesale & 0.157 & 1.602 & 0.157 & 1.528 & \textbf{\underline{0.148}} & \textbf{\underline{1.402}} & 0.157 & 1.528 & \textbf{\underline{0.148}} & \textbf{\underline{1.402}} \\
Yeast & \textbf{0.059} & 1.951 & \textbf{0.059} & \textbf{1.938} & \textbf{0.059} & 1.973 & \underline{0.091} & \underline{2.802} & \underline{0.112} & \underline{4.726} \\
\bottomrule
\addlinespace[\belowrulesep]
\multicolumn{11}{p{0.966\textwidth}}{\footnotesize{Average results of 10-fold cross validation. Significantly different from Multi-class Raw is indicated with underlining. Best performing scenario with each classifier is indicated with boldface font.}}
\end{tabularx}
\end{table}

Breaking up the multi-class problem into one-vs-rest binary problems and combining the results by normalization was able to improve calibration of naive Bayes even without calibrating the binary classifier probabilities on almost all data sets. The same was not true for the all pairs approach that performs worse on some and achieves approximately the same level of performance as uncalibrated multi-class classification on some data sets. Calibrating the binary naive Bayes classifiers in the one-vs-rest approach was able to improve the error metrics on ten of the twelve data sets compared to both uncalibrated multi-class and uncalibrated one-vs-rest scenarios. One exception to this was on the Waveform data set where LL was not significantly different from the uncalibrated one-vs-rest scenario even though MSE was. Calibration did, however, improve both MSE and LL on that data set compared to uncalibrated multi-class classification.

Calibrating the binary naive Bayes classifiers in the all pairs approach improved calibration on seven of the twelve data sets compared to uncalibrated multi-class classification. On two data sets MSE increased while LL decreased and on one of the data sets the treatment increased both MSE and LL.

Overall the one-vs-rest approach with DGG + ENIR calibration coupled with normalization was the best performing calibration scenario for naive Bayes. One-vs-rest calibration performed better than all pairs on five data sets, there was no statistically significant difference on six data sets, and all pairs was better on one data set.

With the random forest classifier, breaking up the multi-class problem into binary problems increased calibration error metrics on four data sets with the one-vs-rest approach and on eight data sets with the all pairs approach. After calibrating the binary problems, calibration improved on six data sets with the one-vs-rest approach and on five data sets with the all pairs approach compared to the corresponding uncalibrated scenario. Compared to the uncalibrated multi-class scenario, calibration performance with the one-vs-rest approach improved on four data sets while being similar on the other eight data sets. The calibrated all pairs was able to improve calibration only on three data sets, was neutral on two data sets, and decreased calibration performance on seven data sets compared to the uncalibrated multi-class scenario.

As with naive Bayes, the one-vs-rest approach fared better than the all pairs approach overall. On seven data sets the one-vs-rest approach did better than the all pairs approach, on four data sets there was no difference, and on one data set the all pairs approach resulted in lower calibration error.

Average computation times for training and calibrating the classifiers were recorded and the results are shown in Table \ref{tab:times}. For the one-vs-rest and the all pairs approaches the calibration times are presented as time consumed for each binary problem to make the numbers comparable when taking into account the number of binary problems on each data set. Naive Bayes was extremely fast to train and although breaking up the classification problem into several binary problems increased the computation times this increase was negligible in practice.

\begin{table}
\centering
\caption{Computation times of building and calibrating the classifiers averaged over the data sets. For One-vs-rest and All Pairs the times are shown as time per binary problem.}
\label{tab:times}       
\begin{tabular}{lll}
\toprule
Scenario & Model & Calibration  \\
\midrule
NB Multi-class & 0.009s & - \\
NB One-vs-rest & 0.052s & 4.36s \\
NB All Pairs & 0.112s & 4.62s \\
\midrule
RF Multi-class & 4.15s & - \\
RF One-vs-rest & 29.5s & 5.13s \\
RF All Pairs & 18.2s & 4.67s \\
\bottomrule
\end{tabular}
\end{table}

For random forest, too, the multi-class classifier was clearly faster to train than either the all pairs or the one-vs-rest. The all pairs classifier was, however, clearly faster to train than the one-vs-rest classifier but with such small data sets this difference is still not very meaningful in practice.

DGG data generation and ENIR calibration took approximately the same time for each binary problem for both the one-vs-rest and the all pairs approaches as the number of generated calibration data points is the same in both approaches. What was a bit surprising was that there was no difference in calibration times, per binary problem, between the classifiers. The overall calibration time then depends mostly on the number of binary problems.

\section{Discussion}
\label{sec:discussion}

Naive Bayes is known to be poorly calibrated because its assumptions about feature independence rarely hold. It is not a big surprise that calibration improves its performance but it is surprising that using the one-vs-rest approach can improve its calibration even without calibrating the binary classifiers. Calibrating the binary naive Bayes classifiers works for both one-vs-rest and all pairs approaches. The calibrated one-vs-rest approach seems to be better suited for naive Bayes than the all pairs and the difference is often statistically significant.

Random forest classifier is not as poorly calibrated as naive Bayes but has still been shown to improve with calibration on some binary problems even with small data sets by using DGG for generating the calibration data set. It is clear from our experiments that the one-vs-rest approach works better with random forest than the all pairs approach does. As the all pairs approach actually decreases calibration performance on some data sets, especially if the number of classes is high, the one-vs-rest is the recommended approach for random forest.

Computation time grew linearly as a function of the number of binary problems because the complexity of DGG data generation depends mainly on the amount of data to be generated which was held constant for each scenario. This indicates that as the number of classes grows so does the calibration time. This might become more of an issue with the all pairs approach than with the one-vs-rest approach. However, the training times for calibration were only a few seconds per binary problem while the prediction times are negligible. In addition, parallel implementation would be trivial to implement which would decrease computation time considerably.

Comparison of the proposed method with calibration approaches that can directly calibrate multi-class probabilities is left for future work.

\section{Conclusions}
\label{sec:conclusions}

Data Generation and Grouping with IR or ENIR calibration can be generalized to multi-class problems as we have shown in this work using ENIR calibration. Using our proposed approach, calibration error can be decreased on many classification problems as demonstrated by our experiments. This is an important finding as traditional calibration algorithms perform poorly on small data sets and not all classification problems are binary. DGG data generation adds computational complexity which grows linearly as a function of binary problems. As the number of binary problems grows more rapidly on the all pairs approach, the one-vs-rest approach has an advantage as the number of classes grows. More importantly, the one-vs-rest approach performs better than the all pairs approach in many cases and did not increase calibration error on any of the tested data sets whereas the all pairs approach does on some of the data sets. The computation times for training the calibration algorithm were merely seconds per binary problem on the tested data sets which is not something that would discourage the usage of this algorithm if good calibration is needed, especially with a parallel implementation.

\begin{acknowledgements}
The authors would like to thank Infotech Oulu, Jenny and Antti Wihuri Foundation, Tauno T\"{o}nning Foundation, and Walter Ahlstr{\"o}m Foundation for financial support of this work.
\end{acknowledgements}

\bibliographystyle{unsrt}
\bibliography{multiclass_calibration}






\end{document}